%% file: main.tex
\documentclass{article} 
\usepackage{iclr2018_conference_blanc,times}
\usepackage{hyperref}
\usepackage{url}
\usepackage{adjustbox}
\usepackage{algorithm}
\usepackage{algorithmic}
\usepackage{amsmath}
\usepackage{graphicx} 
\usepackage{subfig}
\usepackage{tikz}
\usetikzlibrary{shapes.geometric}
\usetikzlibrary{shapes.arrows}
\usetikzlibrary{positioning}

\usepackage{booktabs}

\title{Learning Independent Features with Adversarial Nets for Non-linear ICA}

\author{Phil\'{e}mon Brakel \& Yoshua Bengio\\
MILA, Universit\'{e} de Montr\'{e}al\\
Montr\'{e}al, Canada\\
\texttt{\{philemon.brakel,yoshua.bengio\}@umontreal.ca} \\
}

\iclrfinalcopy 

\begin{document}

\maketitle

\begin{abstract}
Reliable measures of statistical dependence could be useful tools for
learning independent features and performing tasks like source separation using
Independent Component Analysis (ICA).
Unfortunately, many of such measures, like the mutual information, are hard to
estimate and optimize directly. 
We propose to learn independent features with adversarial
objectives \citep{goodfellow2014gan,arjovsky2017wasserstein,huszar2016alternative} which optimize such measures implicitly.
These objectives compare samples from the joint distribution and the product of the marginals
without the need to compute any probability densities. We also propose two methods for
obtaining samples from the product of the marginals using either a simple
resampling trick or a separate parametric distribution.
Our experiments show that this strategy can easily be applied to
different types of model architectures and solve both linear and
non-linear ICA problems.\footnote{A preliminary version of this work was presented at the ICML 2017 workshop on implicit models.}
\end{abstract}

\section{Introduction}

How to extract statistically independent components from data has been
thoroughly investigated in fields like machine learning, statistics and signal
processing under the name Independent Component Analysis (ICA;
\citealt{hyvarinen2004ica}).
ICA has been especially popular as a method for blind source separation (BSS)
and its application domains include medical signal analysis (e.g., EEG and
ECG), audio source separation and image processing (see
\citealt{naik2011overview} for a comprehensive overview of ICA applications).
Various methods for learning ICA models
have been proposed and the method has been extended to mixing models that go
beyond the original linearity and iid assumptions \citep{almeida2003misep,hyvarinen2017nonlinear}.

Many learning algorithms for linear and non-linear Independent Component Analysis are in some way based on a
minimization of the mutual information (MI) or similar measures which compare a joint distribution with the product of its marginals.
A popular example of this is the \emph{Infomax} method \citep{bell1995information}.
Unfortunately, it is often hard to measure and optimize
the MI directly, especially in higher dimensions.
This is why Infomax only minimizes the mutual information indirectly by maximizing the joint entropy instead.
While there is some work on estimators for mutual information and independence based on, for
example, non-parametric methods
\citep{kraskov2004estimating,gretton2005kernel}, it is typically not
straightforward to employ such measures as optimization criteria.

Recently, the framework of Generative Adversarial Networks (GANs) was proposed
for learning generative models \citep{goodfellow2014gan} and matching distributions. GAN training can be seen as
approximate minimization of the Jensen-Shannon divergence
between two distributions without the need to compute densities. Other recent work extended this interpretation of
GANs to other divergences and distances between distributions
\citep{arjovsky2017wasserstein,hjelm2017,mao2016least}.
While most work on GANs applies this matching of distributions in the context
of generative modelling, some recent work has extended these ideas to learn
features which are invariant to different domains \citep{ganin2016domain} or
noise conditions \citep{serdyuk2016invariant}.

We show how the GAN framework allows us to define new objectives for learning statistically independent features.
The gist of the idea is to use adversarial training to train some joint
distribution to produce samples which become indistinguishable from samples of
the product of its marginals.
Our empirical work shows that auto-encoder type models which are trained to optimize our independence objectives
can solve both linear and non-linear ICA problems with different numbers of sources and observations.

\section{Background}
The mutual information of two stochastic variables $Z_1$ and $Z_2$ corresponds to the
Kullback-Leibler (KL) divergence between their
joint density and the product of the marginal densities:
\begin{equation}
    I(Z_1, Z_2)=\int\int p(z_1,z_2) \log \frac{p(z_1,z_2)}{p(z_1)p(z_2)}\mathrm{d}z_1 z_2.
    \label{eq:mi}
\end{equation}
We will often write densities like $p(Z_1=z_1)$ as $p(z_1)$ to save space.
The MI is zero if and only if all the variables are mutually independent.
One benefit of working with MI as a measure of dependence/independence, is that
it can easily be related to other information theoretical quantities, like for
example the entropies of the distributions involved.
Another nice property of the mutual information is that, unlike differential entropy, it is invariant under
reparametrizations of the marginal variables \citep{kraskov2004estimating}. This means that if two functions $f$
and $g$ are homeomorphisms, the $I(Z_1, Z_2)=I(f(Z_1),g(Z_2))$.
Unfortunately, the mutual information is often hard to compute directly, especially without access
to the densities of the joint and marginal distributions.

Generative Adversarial Networks (GANs; \citealt{goodfellow2014gan}) provide a
framework for matching distributions without the need to compute densities.
During training, two neural networks are involved: the \emph{generator} and the \emph{discriminator}. The
generator is a function $G(\cdot)$ which maps samples from a known
distribution (e.g., the unit variance multivariate normal distribution) to
points that live in the same space as the samples of the data set.
The discriminator is a classifier $D(\cdot)$ which is trained to separate the
`fake' samples from the generator from the `real' samples in the data set.
The parameters of the generator are optimized to `fool' the
discriminator and maximize its loss function using gradient information
propagated through the samples.\footnote{While this restricts the use of GANs
to continuous distributions, methods for discrete distributions have been
proposed as well \citep{hjelm2017}.} In the original formulation of
the GAN framework, the discriminator is optimized using the cross-entropy loss. The full definition of the GAN learning objective is given by
\begin{equation}
\label{eq:gan}
\min_{G} \max_{D} E_{\text{data}}[\log D(\mathbf{x})] + E_{\text{gen}}[\log(1-D(\mathbf{y}))],
\end{equation}
where $E_{\text{data}}$ and $E_{\text{gen}}$ denote expectations
with respect to the data and generator distributions.

Since we will evaluate our models on ICA source separation problems, we will
describe this setting in a bit more detail as well.
The original linear ICA model assumes that some observed multivariate signal
$\mathbf{x}$ can be modelled as
\begin{equation}
    \mathbf{x} = \mathbf{As},
\end{equation}
where $\mathbf{A}$ is a linear transformation and 
$\mathbf{s}$ is a set of mutually independent \emph{source} signals which all have non-Gaussian
distributions.
Given the observations $\mathbf{x}$, the goal is to retrieve the source signals
$\mathbf{s}$. When the source signals are indeed non-Gaussian (or there is at
least no more than one Gaussian source), the matrix $\mathbf{A}$ is of full
column rank and the number of observations is at least as large as the number
of sources, linear ICA is guaranteed to be identifiable
\citep{comon1994independent} up to a permutation and rescaling of the sources.
When the mixing of the signals is not linear, identifiability cannot be
guaranteed in general. However, under certain circumstances, some specific
types of non-linear mixtures like post non-linear mixtures (PNL) can still be
separated \citep{taleb1999source}.

\section{Minimizing and Measuring Dependence}
For the moment, assume that we have access to samples from both the joint distribution
$p(\mathbf{z})$ and the product of the marginals $\prod_i p(z_i)$. We now want
to measure how dependent/independent the individual variables of the joint distribution
are without measuring any densities. As pointed out by \citet{arjovsky2017wasserstein}, 
the earth mover's distance between two distributions $q$ and $r$ can, under certain conditions, be approximated by
letting $f(\cdot)$ be a neural network and solving the following optimization
problem:
\begin{equation}
    \max_{\|f\|_L \le 1} E_{\mathbf{z}\sim q(\mathbf{z})}[f(\mathbf{z})] - E_{\mathbf{z}\sim r(\mathbf{z})}[f(\mathbf{z})].
    \label{eq:wasserstein}
\end{equation}
If we substitute $q$ and $r$ for $p(\mathbf{z})$ and $\prod_i p(z_i)$,
respectively, we can consider Equation \ref{eq:wasserstein} to be a measure of
dependence for $p(\mathbf{z})$ that can serve as an alternative to the mutual information.
Now if it is also possible to backpropagate gradients through the samples with
respect to the parameters of the distribution, we can use these to minimize
Equation \ref{eq:wasserstein} and make the variables more independent. Similarly, the
standard GAN objective can be used to approximately minimize the JS-divergence between the
joint and marginal distributions instead.
While we focussed on learning independent features and the measuring of
dependence is not the subject of the research in this paper, we think that
the adversarial networks framework may provide useful tools for this as well.

Finally, as shown in a blog post \citep{huszar2016alternative}, the standard
GAN objective can also be adapted to approximately optimize the KL-divergence.
This objective is obviously an interesting
case because it results in an approximate optimization of the mutual
information itself but in preliminary experiments we found it harder to
optimize than the more conventional GAN objectives.

\subsection{Obtaining the Samples}
So far, we assumed that we had access to both samples from the joint
distribution and from the product of the marginals.
To obtain approximate samples from the product of the marginals, we propose to
either \emph{resample} the values of samples from the joint distribution or to train a separate
generator network with certain architectural constraints.

Given a sample $(z_1,\dots,z_M)^{\mathsf{T}}$
of some joint distribution $p(z_1,\dots,z_M)$,
a sample of the marginal distribution $p(z_1)$ can be obtained by simply
discarding all the other variables from the joint sample. To obtain samples
from the complete product $\prod^{M}_{i=1} p(z_i)$, the same method can
be used by taking $M$ samples from the joint distribution and making sure that
each of the $M$ dimensions from the new factorized sample is taken from a different
sample of the joint. In other words, given $K$ joint samples where $K\ge M$, one can
randomly choose $M$ integers from $\{1,\dots,N\}$ without replacement and use
them to select the elements of the sample from the factorized distribution.
When using sampling \emph{with} replacement, a second sample obtained in this way from the same batch of joint
samples would not be truly independent of the first.
We argue that this is not a big problem in practice as long as one ensures that
the batches are large enough and randomly chosen.
So step by step, we perform the following procedure to obtain samples from the product of the marginal distributions using resampling:
\begin{enumerate}
    \item Sample a batch $\mathbf{Z}$ of $K$ samples from the joint distribution.
    \item Sample a vector $\mathbf{u}$ of $M$ indices uniformly from $\{1,\cdots,K\}$
    \item Construct a resampled vector $\mathbf{\hat{z}}$ where $\hat{z}_i=\mathbf{Z}_{u_{i}i}$.
    \item Repeat step 3 to obtain a matrix of resampled values $\mathbf{\hat{Z}}$.
\end{enumerate}
One could experiment with different numbers of resampled vectors per batch of
data but in our experiments $\mathbf{Z}$ and $\mathbf{\hat{Z}}$ will always have
the same shape.

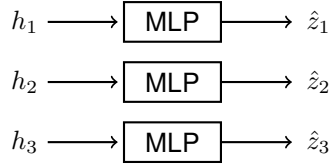
\begin{figure}
    \centering
\begin{tikzpicture} [
    auto,
    action/.style    = { rectangle, draw=black, thick, 
                        fill=none, text width=6em, text centered,
                        rounded corners, minimum height=2em },
    block/.style    = { rectangle, draw=black, thick, 
                        fill=none, text width=3em, text centered,
                        minimum height=1.5em },
    bigblock/.style    = { rectangle, draw=black, thick, 
                        fill=none, text width=6em, text centered,
                        minimum height=2cm },
    line/.style     = { draw, thick, ->, shorten >=2pt },
  ]
\node [text centered] (h2) {$h_2$};
\node [text centered, above = 15pt of h2.center] (h1) {$h_1$};
\node [text centered, below = 15pt of h2.center] (h3) {$h_3$};
\node [block, right = of h1] (mlp1) {\textsf{MLP}};
\node [block, right = of h2] (mlp2) {\textsf{MLP}};
\node [block, right = of h3] (mlp3) {\textsf{MLP}};
\node [text centered, right = of mlp1] (z1) {$\hat{z}_1$};
\node [text centered, right = of mlp2] (z2) {$\hat{z}_2$};
\node [text centered, right = of mlp3] (z3) {$\hat{z}_3$};
  \begin{scope} [every path/.style=line]
    \path (h1)   --    (mlp1);
    \path (h2)   --    (mlp2);
    \path (h3)   --    (mlp3);
    \path (mlp1)   --    (z1);
    \path (mlp2)   --    (z2);
    \path (mlp3)   --    (z3);
  \end{scope}
\end{tikzpicture}
\caption{Example of a directed latent variable generator architecture which only supports factorized distributions.}
\label{fig:generator}
\end{figure}

Another way to simulate the product of marginal distributions is by using a
separate generator network which is trained to optimize the same objective as
the generator of the joint distribution.
By sampling independent latent variables and transforming each of them with a
separate multi-layer perceptron (see Figure~\ref{fig:generator}), this generator should be able to learn to
approximate the joint distribution with a factorized distribution without imposing a specific prior.
While it may be more difficult to learn the marginal distributions explicitly,
it could in some situations be useful to have an explicit model for this
distribution available after training for further analysis or if the goal is to
build a generative model.
This approach may be especially useful when the data are not iid (like in time
series) and one doesn't want to ignore the inter-sample dependencies like the resampling method does.

\section{Adversarial Non-linear Independent Components Analysis}
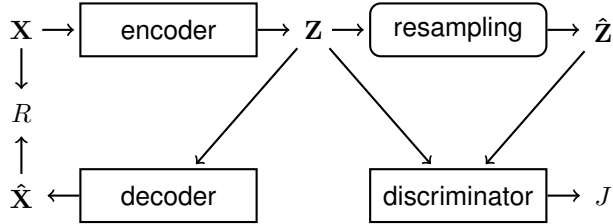
\begin{figure}
    \centering
\begin{tikzpicture} [
    auto,
    action/.style    = { rectangle, draw=black, thick, 
                        fill=none, text width=6em, text centered,
                        rounded corners, minimum height=2em },
    block/.style    = { rectangle, draw=black, thick, 
                        fill=none, text width=6em, text centered,
                        minimum height=2em },
    line/.style     = { draw, thick, ->, shorten >=2pt },
  ]
  \matrix [column sep=5mm, row sep=5mm] {
                    \node [text centered] (x) {$\mathbf{X}$}; &
                    \node [block] (encoder) {\textsf{encoder}}; &
                    \node [text centered] (z) {$\mathbf{Z}$}; &
                    \node [action] (resampling) {\textsf{resampling}}; &
                    \node [text centered] (zhat) {$\mathbf{\hat{Z}}$}; & \\
                    \node [text centered] (R) {$R$}; \\
                    \node [text centered] (xhat) {$\mathbf{\hat{X}}$}; &
                    \node [block] (decoder) {\textsf{decoder}}; & &
                    \node [block] (discriminator) {\textsf{discriminator}}; &
                    \node [text centered] (J) {$J$}; \\
  };
  \begin{scope} [every path/.style=line]
    \path (x)        --    (encoder);
    \path (x)        --    (R);
    \path (xhat)        --    (R);
    \path (encoder)        --    (z);
    \path (z)        --    (decoder);
    \path (decoder)        --    (xhat);
    \path (z)        --    (resampling);
    \path (z)        --    (discriminator);
    \path (resampling)        --    (zhat);
    \path (zhat)        --    (discriminator);
    \path (discriminator)        --    (J);
  \end{scope}
\end{tikzpicture}
\caption{Schematic representation of the entire system for learning non-linear ICA. Specific functional shapes can be enforced by choosing a suitable decoder architecture.}
\label{fig:system}
\end{figure}

As a practical application of the ideas described above, we will now develop a system
for learning independent components.
The goal of the system is to learn an \emph{encoder} network $F(\mathbf{x})$ which maps
data/signals to informative features $\mathbf{z}$ which are mutually independent.
We will use an adversarial objective to achieve this in the manner described above.
However, enforcing independence by itself does not guarantee that the mapping from the
observed signals $\mathbf{x}$ to the predicted sources $\mathbf{z}$ is
informative about the input.
To enforce this, we add a \emph{decoder} which
tries to reconstruct the data from the predicted features as was done by
\citet{schmidhuber1992learning}.
Figure~\ref{fig:system} shows a schematic representation of the training setup in its entirety.

\begin{algorithm}
    \caption{Adversarial Non-linear ICA train loop}
    \label{alg:anica}
    \begin{algorithmic}
        \INPUT data $\mathcal{X}$, encoder $F$, decoder $V$, discriminator $D$
        \WHILE{Not converged}
        \STATE sample a batch $\mathbf{X}$ of $N$ data (column) vectors from $\mathcal{X}$
        \STATE $\mathbf{Z}\gets F(\mathbf{X})$  // apply encoder
        \STATE $\mathbf{\hat{X}}\gets V(\mathbf{Z})$  // apply decoder
        \FOR{$j\in \{1,\cdots,N\}$}
            \FOR{$i\in \{1,\cdots,M\}$}
                \STATE $k\gets \text{Uniform}(\{1,\cdots,N\})$  // sample col.\ index
                \STATE $\hat{Z}_{ij}\gets X_{ik}$
            \ENDFOR
        \ENDFOR
        \STATE $J=\log(D(\mathbf{Z})) + \log(1-D(\mathbf{\hat{Z}}))$
        \STATE Update $D$ to maximize $J$
        \STATE $R=\|\mathbf{X}-\mathbf{\hat{X}}\|_{2,1}$
        \STATE Update $F$ and $V$ to minimize $J + \lambda R$
        \ENDWHILE
    \end{algorithmic}
\end{algorithm}

Given the encoder, the decoder, the discriminator, samples from the data, the joint
distribution, and the product of the marginals, we can now compute the GAN objective
from Equation \ref{eq:gan} (or Equation \ref{eq:wasserstein}, assuming the Lipschitz
constraint is also enforced) and add the reconstruction objective to it.
The full procedure of the resampling version of our setup using the standard
GAN objective is given by Algorithm~\ref{alg:anica}.
When a separate generator is used, $\mathbf{\hat{Z}}$ is sampled from it
directly and its parameters participate in the minimization of objective $J$.

Finally, we found that it is important to normalize the features before permuting them and presenting them to the discriminator.
This prevents them both from going to zero and from growing indefinitely in
magnitude, potentially causing the discriminator to fail because it cannot keep
up with the overall changes of the feature distribution.
We also used these normalized features as
input for the decoder, followed by an element-wise rescaling using trainable
parameters, similar to what is done in batch normalization
\citep{ioffe2015batch}.
Without normalization of the decoder inputs, the models would sometimes get
stuck in degenerate solutions.

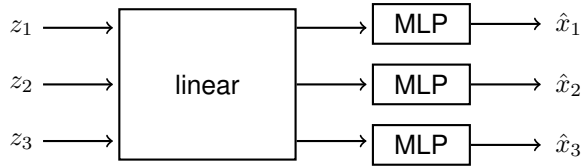
\begin{figure}
    \centering
\begin{tikzpicture} [
    auto,
    action/.style    = { rectangle, draw=black, thick, 
                        fill=none, text width=6em, text centered,
                        rounded corners, minimum height=2em },
    block/.style    = { rectangle, draw=black, thick, 
                        fill=none, text width=3em, text centered,
                        minimum height=1.5em },
    bigblock/.style    = { rectangle, draw=black, thick, 
                        fill=none, text width=6em, text centered,
                        minimum height=2cm },
    line/.style     = { draw, thick, ->, shorten >=2pt },
  ]
\node [bigblock] (linear) {\textsf{linear}};
\node [text centered, left = of linear] (z2) {$z_2$};
\node [text centered, above = 15pt of z2.center] (z1) {$z_1$};
\node [text centered, below = 15pt of z2.center] (z3) {$z_3$};
\node [block, right = of linear] (mlp2) {\textsf{MLP}};
\node [block, above = 15pt of mlp2.center] (mlp1) {\textsf{MLP}};
\node [block, below = 15pt of mlp2.center] (mlp3) {\textsf{MLP}};
\node [text centered, right = of mlp1] (x1) {$\hat{x}_1$};
\node [text centered, right = of mlp2] (x2) {$\hat{x}_2$};
\node [text centered, right = of mlp3] (x3) {$\hat{x}_3$};
  \begin{scope} [every path/.style=line]
    \path (z1.east)        --    (z1 -| linear.west);
    \path (z2.east)        --     (z2 -| linear.west);
    \path (z3.east)        --     (z3 -| linear.west);
    \path (z1 -| linear.east)        --    (z1.east -| mlp1.west);
    \path (z2 -| linear.east)        --    (z2.east -| mlp2.west);
    \path (z3 -| linear.east)        --    (z3.east -| mlp3.west);
    \path (mlp1)   --    (x1);
    \path (mlp2)   --    (x2);
    \path (mlp3)   --    (x3);
  \end{scope}
\end{tikzpicture}
\caption{The decoder architecture used for the PNL experiments. It can only
learn transformations in which a linear transformation is followed by the
application of non-linear scalar functions to each of the dimensions.}
\label{fig:dec}
\end{figure}

\section{Related Work}
Most optimization methods for ICA are either based on non-Gaussianity, like the
popular FastICA algorithm \citep{hyvarinen1997one}, or on minimization of the
mutual information of the extracted source signals, as is implicitly done with
Infomax methods \citep{bell1995information}.
The Infomax ICA algorithm maximizes the joint entropy of the estimated signals.
Given a carefully constructed architecture, the marginal entropies are bounded and
the maximization leads to a minimization of the mutual information.
Infomax has been extended to
non-linear neural network models and the MISEP model
\citep{almeida2003misep} is a successful example of this.
Infomax methods don't need an additional decoder
component to ensure invertibility and there are no sampling methods involved.
Unlike our model however, training involves a computation of the gradient of
the logarithm of the determinant of the jacobian for each data point.
This can be computationally demanding and also requires the number of sources and
mixtures to be equal. Furthermore, the our method provides a way of promoting
independence of features decoupled from maximizing their information.

\label{sec:related}
This work was partially inspired by J\"{u}rgen Schmidhuber's work on the
learning of binary factorial codes \citep{schmidhuber1992learning}. In that
work, an auto-encoder is also combined with an adversarial objective,
but one based on the mutual predictability of the variables rather than separability
from the product of the marginals.
To our knowledge, this method for learning binary codes has not yet been
adapted for continuous variables.

The architectures in our experiments are also similar to Adversarial
Auto-Encoders (AAEs) \citep{makhzani2015adversarial}. In AAEs, the GAN
principle is used to match the distribution at the output of an encoder when fed by the data with some prior as
part of a Variational Autoencoder (VAE) \citep{kingma2013auto} setup. Similar
to in our work, the KL-divergence between two distributions is replaced with
the GAN objective. When a factorized prior is used (as is
typically done), the AAE also learns to produce independent
features. However, the chosen prior also forces the learned features to adhere
to its specific shape and this may be in competition with the independence property.
We actually implemented uniform and normal priors for our model but were not
able to learn signal separation with those.
Another recent related model is InfoGAN \citep{chen2016infogan}.
InfoGAN is a generative GAN model in which the mutual information between some
latent variables and the outputs is maximized. While this also promotes
independence of some of the latent variables, the desired goal is now to provide more control
over the generated samples.

Some of the more successful estimators of mutual information are based on
nearest neighbor methods which compare the relative distances of complete
vectors and individual variables \citep{kraskov2004estimating}.
An estimator of this type has also been used to perform linear blind source
separation using an algorithm in which different rotations of components are
compared with each other \citep{stogbauer2004least}. Unfortunately, this estimator is biased when
variables are far from independent and not differentiable, limiting it's use as
a general optimization criterion.
Other estimators of mutual information and dependence/independence in general are based on kernel methods \citep{gretton2005kernel,gretton2008kernel}.
These methods have been very at linear ICA but have, to our knowledge, not been
evaluated on more general non-linear problems.

Finally, there has been recent work on invertible non-linear mappings that
allows the training of tractable neural network latent variable models which
can be interpreted as non-linear independent component analysis. Examples of
these are the NICE and real-NVP models \citep{dinh2014nice,dinh2016density}.
An important difference with our work is that these models require one to
specify the distribution of the source signals in advance.

\section{Experiments}

We looked at linear mixtures, post non-linear mixtures which are not linear but still separable
and overdetermined general non-linear mixtures which may not be separable.\footnote{Code for training both Anica and PNLMISEP models can be found online: \url{https://github.com/pbrakel/anica}}
ICA extracts the source signals only up to a permutation and scaling.
Therefore, all results are measured by considering all possible pairings of the predicted
signals and the source signals and measuring the average absolute correlation of the best pairing.
We will just refer to this as $\rho_{\text{max}}$ or simply `correlation'.
We will refer to our models with the name `Anica', which is short for
Adversarial Non-linear Independent Component Analysis.

\begin{figure}
    \centering
    \subfloat[Synthetic source signals.\label{subfig:sources}]{
        \includegraphics[scale=.3]{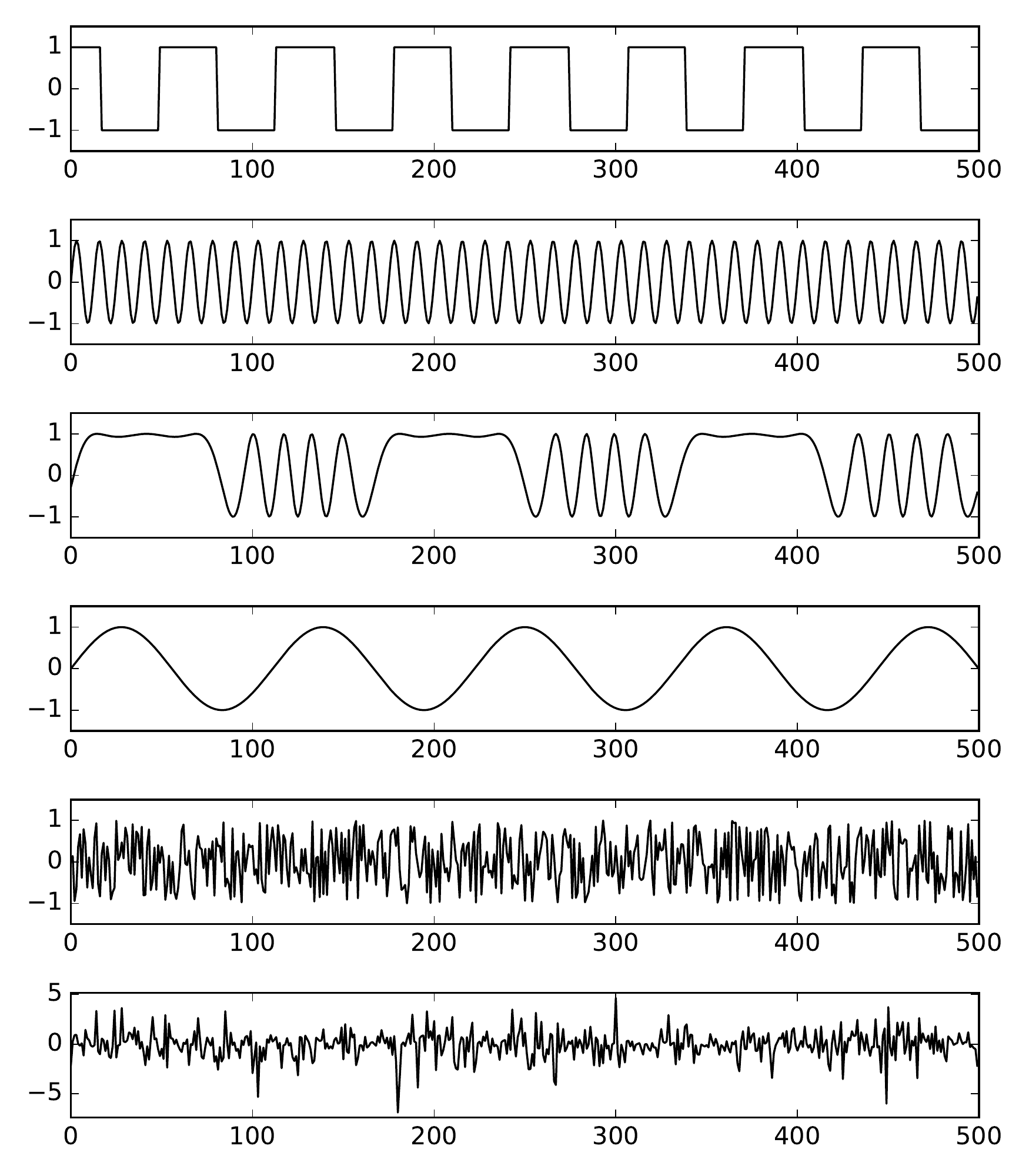}
    }
    \subfloat[Audio source signals.\label{subfig:sourcesaudio}]{
        \includegraphics[scale=.3]{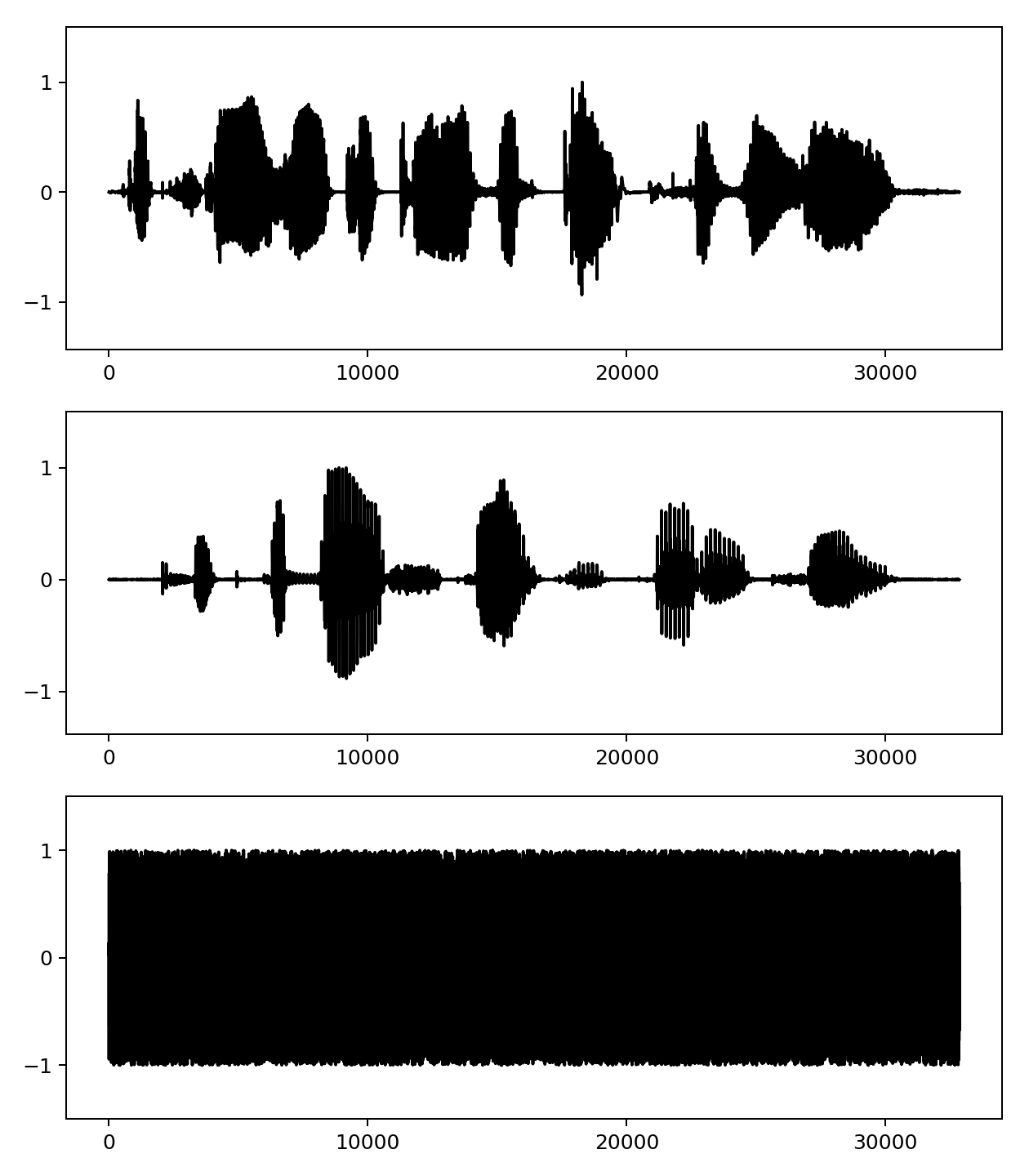}
    }
    \caption{Source signals used in the experiments.}
    \label{fig:sources}
\end{figure}

\paragraph{Source signals}
We used both synthetic signals and actual speech signals as sources for our experiments (see Figure \ref{fig:sources}).
The synthetic source signals were created with the goal to include both sub-gaussian and
super-gaussian distributions, together with some periodic signals for
visualization purposes.
The data set consisted of the first 4000 samples of these signals.\footnote{See
the appendix for more details about the synthetic signals.}
For the audio separation tasks, we used speech recordings
from the 16kHz version of the freely available TSP data set
\citep{kabal2002tsp}. The first source was an utterance from a male speaker
(\texttt{MA02\_04.wav}), the second source an utterance from a female speaker
(\texttt{FA01\_03.wav}), and the third source was uniform noise.
All signals were normalized to have a peak amplitude of 1.
The signals were about two seconds long, which translated to roughly 32k samples.

\paragraph{Linear ICA}
As a first proof of concept, we trained a model in which both the model and the
transformation of the source signals are linear. 
The mixed signals $\mathbf{x}$ were created by forming a matrix $\mathbf{A}$
with elements sampled uniformly from the interval $[-.5,.5]$ and multiplying it
with the source signals $\mathbf{s}$.
Both the encoder and decoder parts of the model were linear
transformations. The discriminator network was a multilayer perceptron with one hidden layer of 64 rectified linear units.

\paragraph{Post non-linear mixtures}
To generate post non-linear mixtures, we used the
same procedure as we used for generating the linear mixture, but followed by a
non-linear function. For the synthetic signals we used the hyperbolic tangent function.
For the audio data we used a different function for each of the three mixtures: $g_1(x)=\tanh(x)$, $g_2(x)=(x + x^3)/2$ and $g_3(x)=e^x$.
We found during preliminary experiments that we
obtained the best results when the encoder, which learns the inverse of the
mixing transformation, is as flexible as possible, while the decoder is
constrained in the types of functions it can learn.
One could also choose a flexible decoder while keeping the encoder constrained
but this didn't seem to work well in practice.
The encoder was a multi-layer perceptron (MLP)
with two hidden layers of rectified linear units (ReLU; \citealt{nair2010rectified}).
The first layer of the decoder was a linear transformation. Subsequently, each output was
processed by a separate small MLP with two layers of 16 hidden ReLU units and a single input
and output. This decoder was chosen to constrain the model to PNL compatible functions.
Note that we did not use any sigmoid functions in our model.
The discriminator network was again multilayer perceptron with one hidden layer of 64 rectified linear units.

\paragraph{Over-determined multi-layer non-linear mixture}
With this task, we illustrate the benefit of our method when there are
more mixture signals than sources for general non-linear mixture problem.
The transformation of the source signals was
$\mathbf{x}=\tanh(\cdot\mathbf{B}\tanh(\mathbf{A}\mathbf{s}))$, where $\mathbf{A}$ and
$\mathbf{B}$ were randomly sampled matrices of $24 \times 6$ and $24 \times 24$
dimensions, respectively.
Both the encoder and decoder for this task were MLPs with two hidden layers of
ReLU units.
The discriminator had two hidden layer with the same number of hidden units as
was chosen for the encoder and decoder networks.
There is no guarantee of identifiability for this task, but the
large number of observations makes it more likely.

\paragraph{Baselines}
For the linear problems, we compared our
results with the FastICA \citep{hyvarinen1997one} implementation from
Scikit-learn
\citep{pedregosa2011scikit} (we report the PNL and MLP results as well just because it's possible).
For the PNL problems, we implemented a version of the MISEP model
\citep{almeida2003misep} with a neural network architecture specifically proposed for these
types of problems \citep{zheng2007misep}.
We also computed $\rho_{\text{max}}$ for the mixed signals.
Unfortunately, we couldn't find a proper baseline for the over-determined MLP
problem.

\subsection{Optimization and hyper-parameter tuning selection}
\label{sec:modelselect}

Quantative evaluation of adversarial networks is still an open problem.
We found that when we measured the sum of the
adversarial loss and the reconstruction loss on held-out data, the model with
the lowest loss was typically not a good model in terms of signal separation.
This can for example happen when the discriminator diverges and the adversarial
loss becomes very low even though the features are not independent.
When one knows how the source signals are supposed to look (or sound) but even
then, this would not be a feasible way to compare numerous models with
different hyper-parameter settings. We found that the reliability of the score,
measured as the standard deviation over multiple experiments with identical
hyper-parameters, turned out to be a much better indicator of signal separation
performance.

For each model, we performed a random search over the number of hidden units in
the MLPs, the learning rate and the scaling of the initial weight matrices of
the separate modules of the model. For each choice of hyper-parameters, we ran
five experiments with different seeds. After discarding diverged models, we
selected the models with the lowest standard deviation in optimization loss on
a held-out set of 500 samples. We report both the average correlation scores of the
model settings selected in this fashion and the ones which were highest on
average in terms of the correlation scores themselves. The latter
represent potential gains in performance if in future work more principled
methods for GAN model selection are developed.
To make our baseline as strong as possible, we performed a similar
hyper-parameter search for the PNLMISEP model to select the number of hidden
units, initial weight scaling and learning rate.
All models were trained for 500000 iterations on batches of 64 samples using
RMSProp \citep{tieleman2012lecture}.

The standard JS-divergence optimizing GAN loss was used for all the hyper-parameter tuning experiment.
We didn't find it necessary to use the commonly used modification of this loss
for preventing the discriminator from saturating. We hypothesize that this
is because the distributions are very similar during early training, unlike the
more conventional GAN problem where one starts by comparing data samples to
noise.
For investigating the convergence behavior we also looked at the results of a
model trained with the Wasserstein GAN loss and gradient penalty \citep{gulrajani2017improved}.

\subsection{Results}
As Table~\ref{tab:results_synth} shows, the linear problems get solved up to
very high precision for the synthetic tasks by all models.
The PNL correlations obtained by the Anica models for the synthetic signals were slightly worse
than of the PNLMISEP baseline. Unfortunately, the model selection procedure also
didn't identify good settings for the Anica-g model and there is a large
discrepancy between the chosen hyper-parameter settings and the ones that lead
to the best correlation scores.
The MLP results on the MLP task were high in general and the scores of the best
performing hyper-parameter settings are on par with those for the PNL task.

On the audio tasks (see Table~\ref{tab:results_audio}), the results for the
linear models were of very high precision but not better than those obtained
with FastICA, unless one would be able to select settings based on the
correlation scores directly. On the PNL task, the resampling based model scored
better than the baseline. The Anica-g model scored worse when the
hyper-parameter selection procedure was used but the score of the best working
settings suggests that it might do similarly well as the resampling model with a
better model selection procedure.
See the appendix for some reconstruction plots of some of the individual models.

\begin{figure}
    \centering
    \subfloat[GAN disc. cost\label{subfig:gandisc}]{
        \includegraphics[scale=.2]{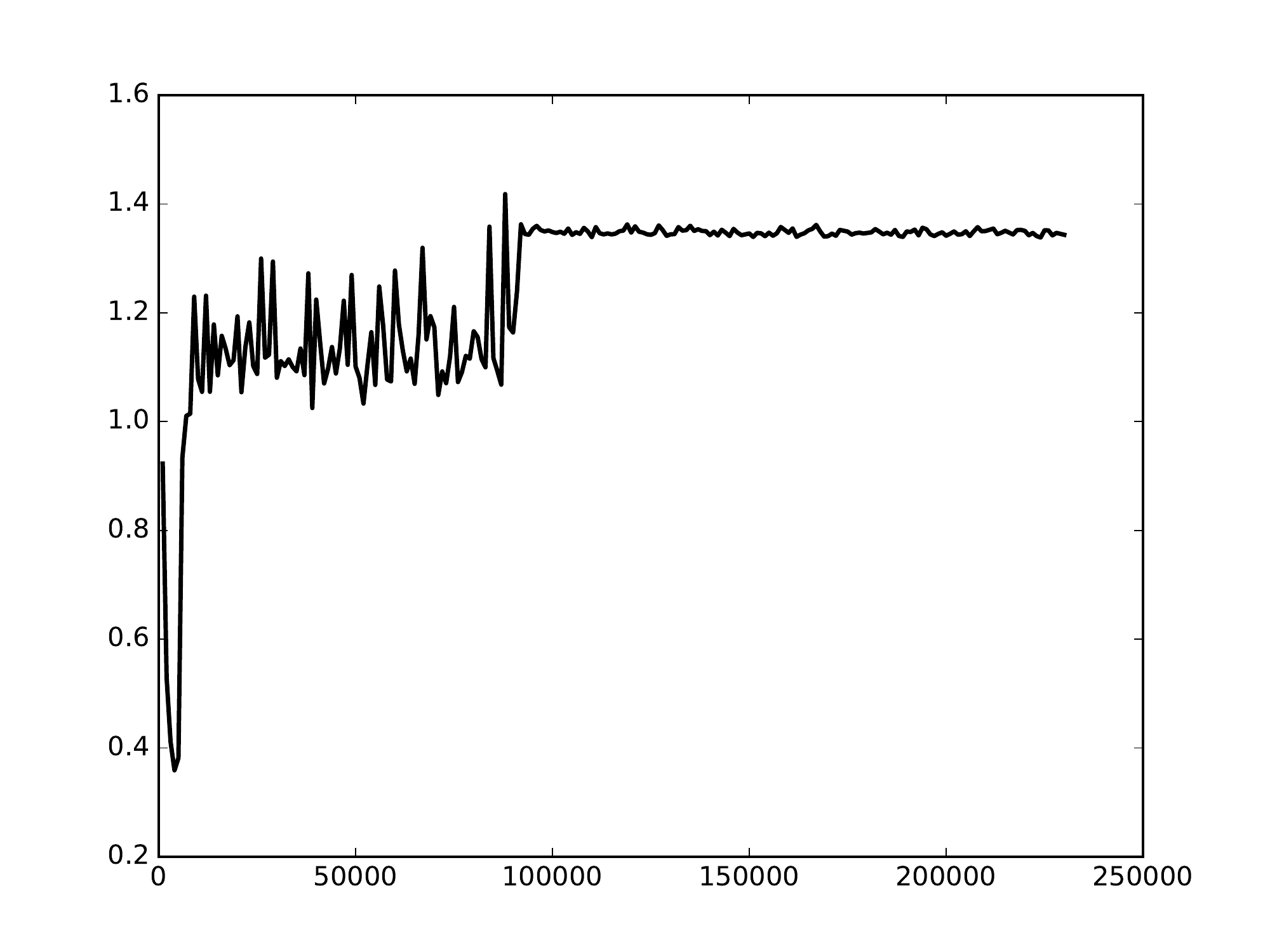}
    }
    \subfloat[WGAN disc. cost\label{subfig:wgandisc}]{
        \includegraphics[scale=.2]{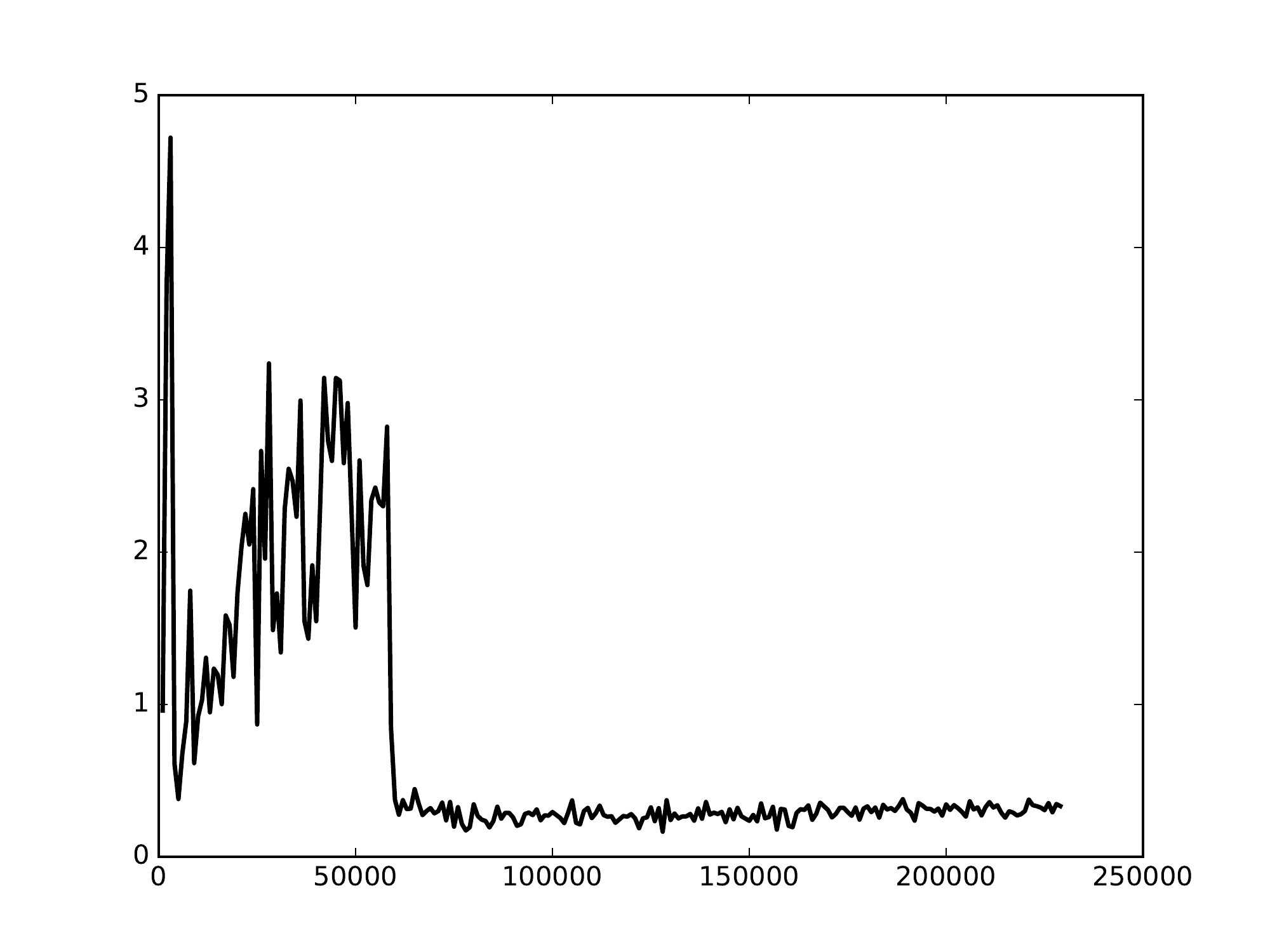}
    }
    \\
    \subfloat[Reconstruction cost\label{subfig:reccost}]{
        \includegraphics[scale=.2]{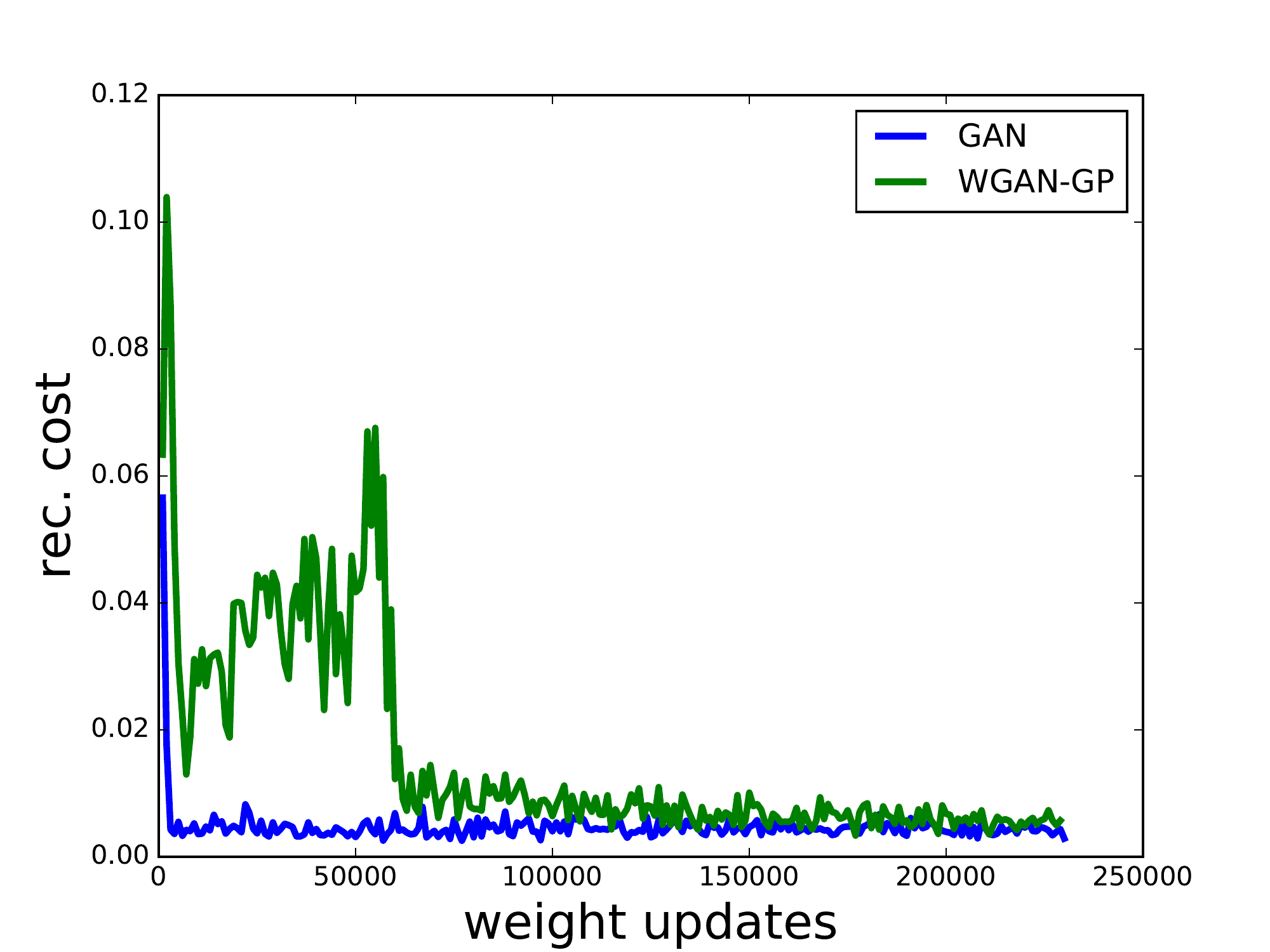}
    }
    \subfloat[Max correlation\label{subfig:maxcorr}]{
        \includegraphics[scale=.2]{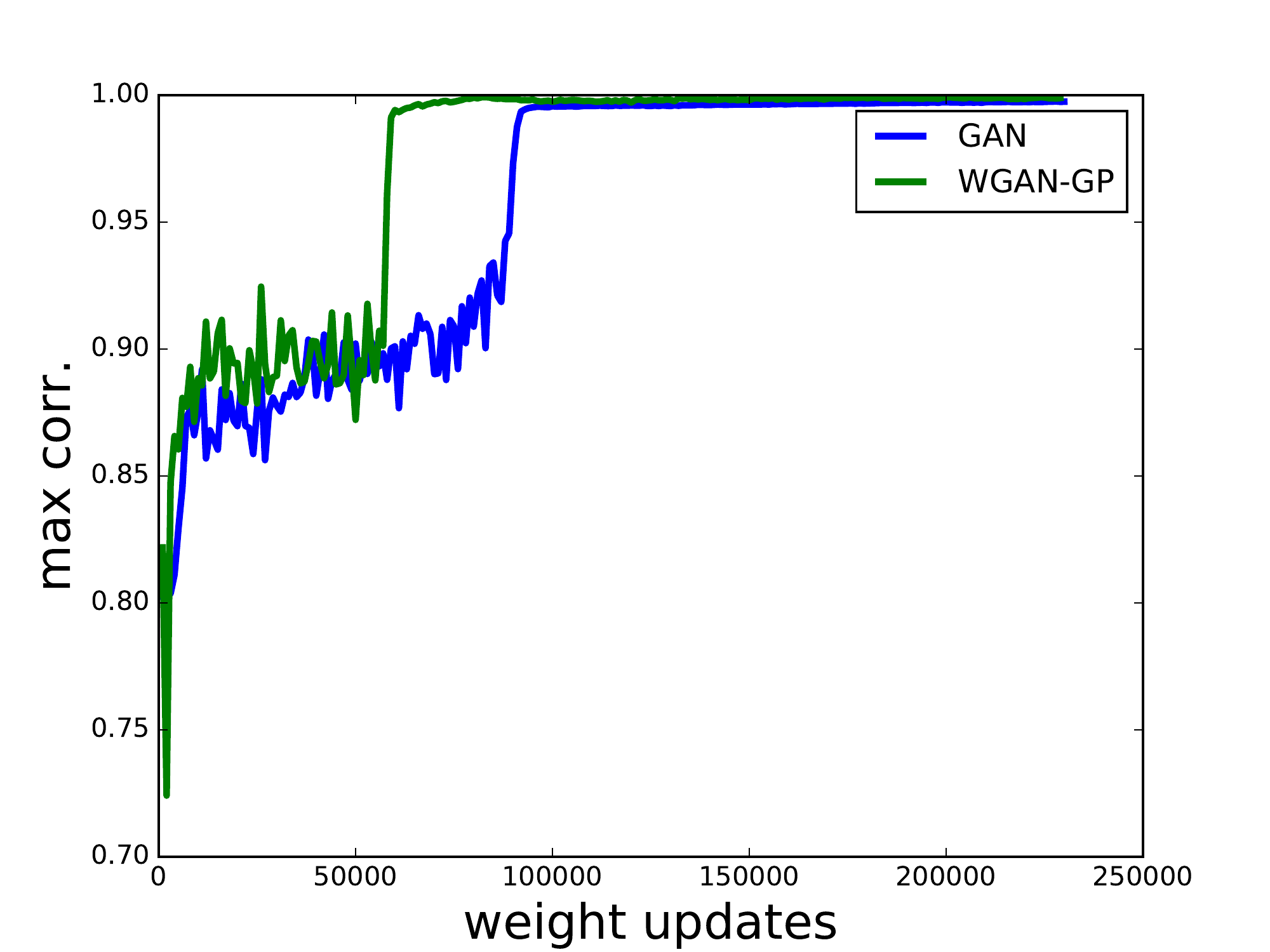}
    }
    \caption{Convergence plots for the linear synthetic source task.}
    \label{fig:convplots}
\end{figure}

To get more insight in the convergence behavior of the models, we plotted
the correlations with the source signals, the discriminator costs and the
reconstruction costs of two linear models for the synthetic signals in
Figure~\ref{fig:convplots}.
For both a GAN and a WGAN version of the resampling-based model, the recognition and
discriminator costs seem to be informative about the convergence of the
correlation scores. However, we also observed situations in which the losses
made a sudden jump after being stuck at a suboptimal value
for quite a while and this might indicate why the consistency of the scores may
be more important than their individual values.

\begin{table}
\caption{Maximum correlation results on all tasks for the synthetic data. A `g' in the suffix
of the model name indicates that a separate generator network was used instead of the resampling
method. Two scores separated by a `/' indicate that the first score was obtained using the model selection described in Section~\ref{sec:modelselect} while the second score is simply the best score \emph{a posteriori}. Parentheses refer contain the standard deviation of the scores multiplied with $10^{-4}$.}
\label{tab:results_synth}
    \begin{center}
        \begin{tabular}{l c c c }
            \toprule
            Method      &Linear                 &PNL                   &MLP                     \\ \midrule
            Anica       &.9987(6.5)/.9994(1.4)  &.9794(53)/.9877(7.9)  &.9667(325)/.9831(16)    \\
            Anica-g     &.9996(1.2)/.9996(1.2)  &.7098(724)/.9802(47)  &.9770(33)/.9856(10.8)   \\
            PNLMISEP    &-                      &.9920(24)              &-                       \\
            FastICA     &.9998                  &.8327                 &.9173                   \\
            Mixed       &.5278                  &.6174                 &-                       \\ \bottomrule
        \end{tabular}
    \end{center}
\end{table}

\begin{table}
\caption{Maximum correlation results on all tasks for the audio data. A `g' in the suffix
of the model name indicates that a separate generator network was used instead of the resampling
method. Two scores separated by a `/' indicate that the first score was obtained using the model selection described in Section~\ref{sec:modelselect} while the second score is simply the best score \emph{a posteriori}. Parentheses refer contain the standard deviation of the scores multiplied with $10^{-4}$.}
\label{tab:results_audio}
    \begin{center}
        \begin{tabular}{l c c}
            \toprule
            Method      &Linear                   &PNL   \\ \midrule
            Anica       &.9996(4.9)/1.0(.1)       &.9929(18)/.9948(12)  \\
            Anica-g     &.9996(3.1)/1.0(.1)       &.9357(671)/.9923(19)     \\
            PNLMISEP    &-                        &.9567(471)        \\
            FastICA     &1.0                      &.8989    \\
            Mixed       &.5338                    &.6550    \\ \bottomrule
        \end{tabular}
    \end{center}
\end{table}

\section{Discussion}

As our results showed, adversarial objectives can successfully be used to learn
independent features in the context of non-linear ICA source separation.
We showed that the methods can be applied to a variety of architectures, work
for signals that are both sub-gaussian and super-gaussian. The method were also
able so separate recordings of human speech.

We found that the generator-based Anica models were a bit harder to optimize and it
was only after systematic hyper-parameter tuning that we were able to get them
to work on the non-linear tasks. This is not too surprising given that one now
has to learn a distribution for the marginal distributions as well. 

Currently, the biggest limitation of our methods is the interpretability of the
optimization objective. That said, we had reasonable success by using the standard deviation
of the scores as a model selection criterion rather than the average of the
scores corresponding to certain hyper-parameter settings.
In the case of identifiable ICA problems and adversarial losses which rely on a
balance between the generator/encoder and the discriminator this makes some intuitive sense.
Perhaps all good models converge to solutions that are alike; where each bad model is bad in its own way.
This performance evaluation issue applies to GANs in general and we
hope that future work on convergence measures for GANs will also improve the
practical applicability of our methods by allowing for more principled model
selection. 

To conclude, our results show that adversarial objectives can be used to maximize
independence and solve linear and
non-linear ICA problems. While the ICA models we implemented are not
always easy to optimize, they seem to work well in practice and can easily be
applied to various different types of architectures and problems.
Future work should be devoted to a more thorough theoretical analysis of
of the proposed methods for minimizing and measuring dependence and how to evaluate them.

\subsubsection*{Acknowledgments}

The authors thank the CHISTERA project M2CR (PCIN-2015-226), Samsung Institute of
Advanced Techonology and CIFAR for their financial support. They also thank Devon Hjelm,
Dzmitry Bahdanau, Ishmael Belghazi, Aaron Courville, Kundan Kumar and Shakir Mohamed for helpful comments.

\bibliography{refs}
\bibliographystyle{iclr2018_conference}

\include{appendix}

\end{document}

%% file: appendix.tex
\newpage
\thispagestyle{empty}
\onecolumn
\section*{Appendix}
\appendix

\section{Synthetic Signals}

The synthetic signals were defined as follows:
\begin{align*}
s_1(t) &= \text{sign}(\cos(310\pi t)), \\
s_2(t) &= \sin(1600\pi t), \\
s_3(t) &= \sin(600\pi t + 6 \cos(120\pi t)), \\
s_4(t) &= \sin(180\pi t), \\
s_5(t) &\sim \text{uniform}(x|[-1, 1]), \\
s_6(t) &\sim \text{laplace}(x|\mu=0,b=1). \\
\end{align*}
The experiments were done using the first 4000 samples with $t$ linearly spaced between $[0, 0.4]$.

\section{Figures}

\begin{figure}[h]
    \centering
    \subfloat[Source signals.\label{subfig:sourcesbig}]{
        \includegraphics[scale=.35]{sources.pdf}
    }
    \subfloat[Anica reconstructions $\rho_{\text{max}}=.997$.\label{subfig:bestlin}]{
        \includegraphics[scale=.35]{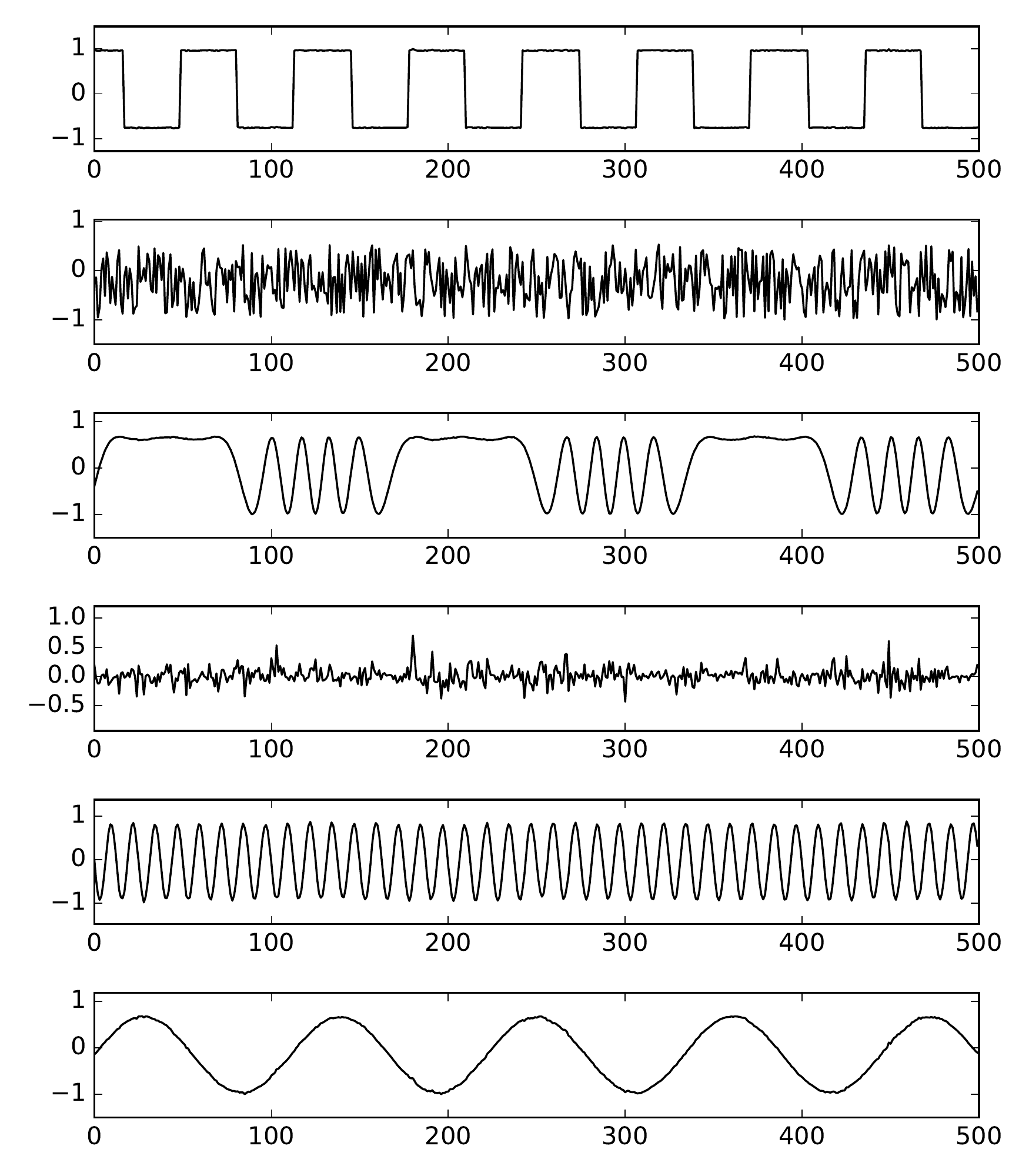}
    }
    \caption{Sources and reconstructions for the linear synthetic source ICA task. The predictions have been
        rescaled to lie within the range $[-1,1]$ for easier comparison with
        the source signals. This causes the laplacian samples to appear scaled
        down. The scores $\rho_{\text{max}}$ represent the maximum absolute
    correlation over all possible permutations of the signals.}
\end{figure}

\begin{figure}
    \centering
    \subfloat[Anica PNL reconstructions $\rho_{\text{max}}=.997$.\label{subfig:anicagpnl}]{
        \includegraphics[scale=.35]{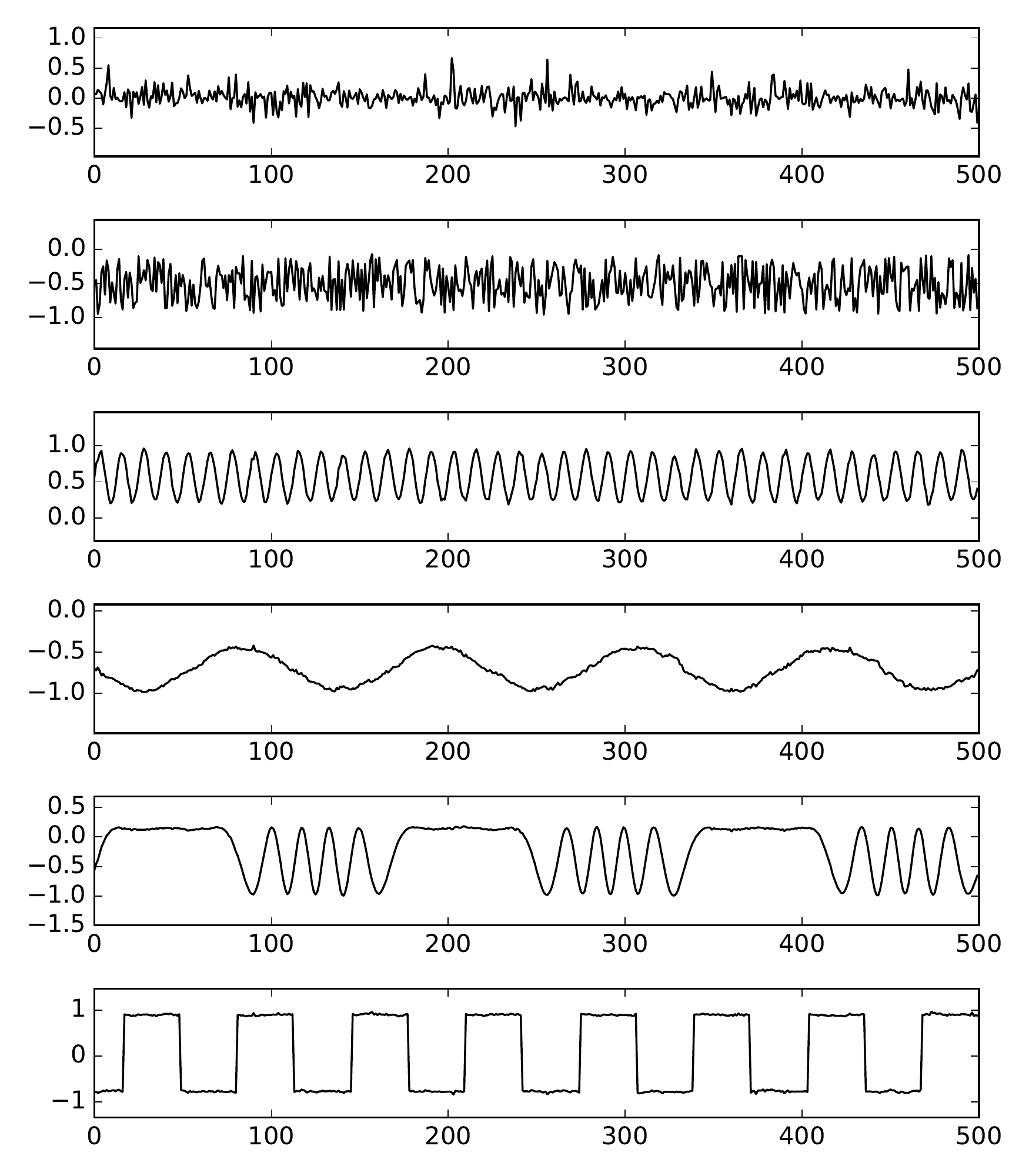}
    }
    \subfloat[Anica MLP reconstructions $\rho_{\text{max}}=.968$.\label{subfig:anicagmlp}]{
        \includegraphics[scale=.35]{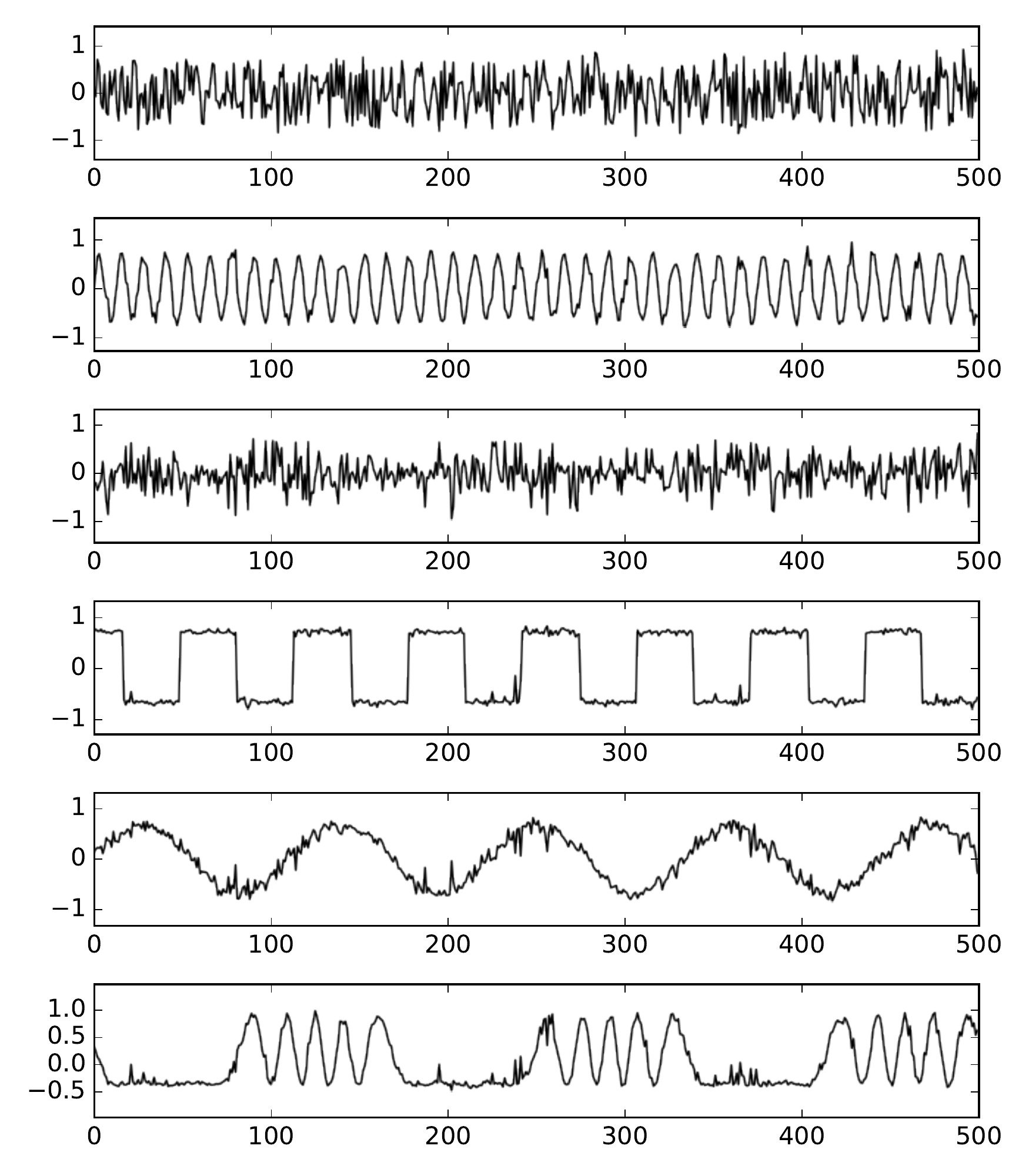}
    }
    \caption{Reconstructions for the post-nonlinear mixture and MLP mixture of the synthetic
    sources.}
\end{figure}

\begin{figure}
    \centering
    \subfloat[Audio source signals.\label{subfig:sourcesaudio2}]{
        \includegraphics[scale=.35]{sourceswav180dpi.png}
    }
    \subfloat[Anica PNL audio reconstructions $\rho_{\text{max}}=.996$.\label{subfig:goodpnlwav}]{
        \includegraphics[scale=.35]{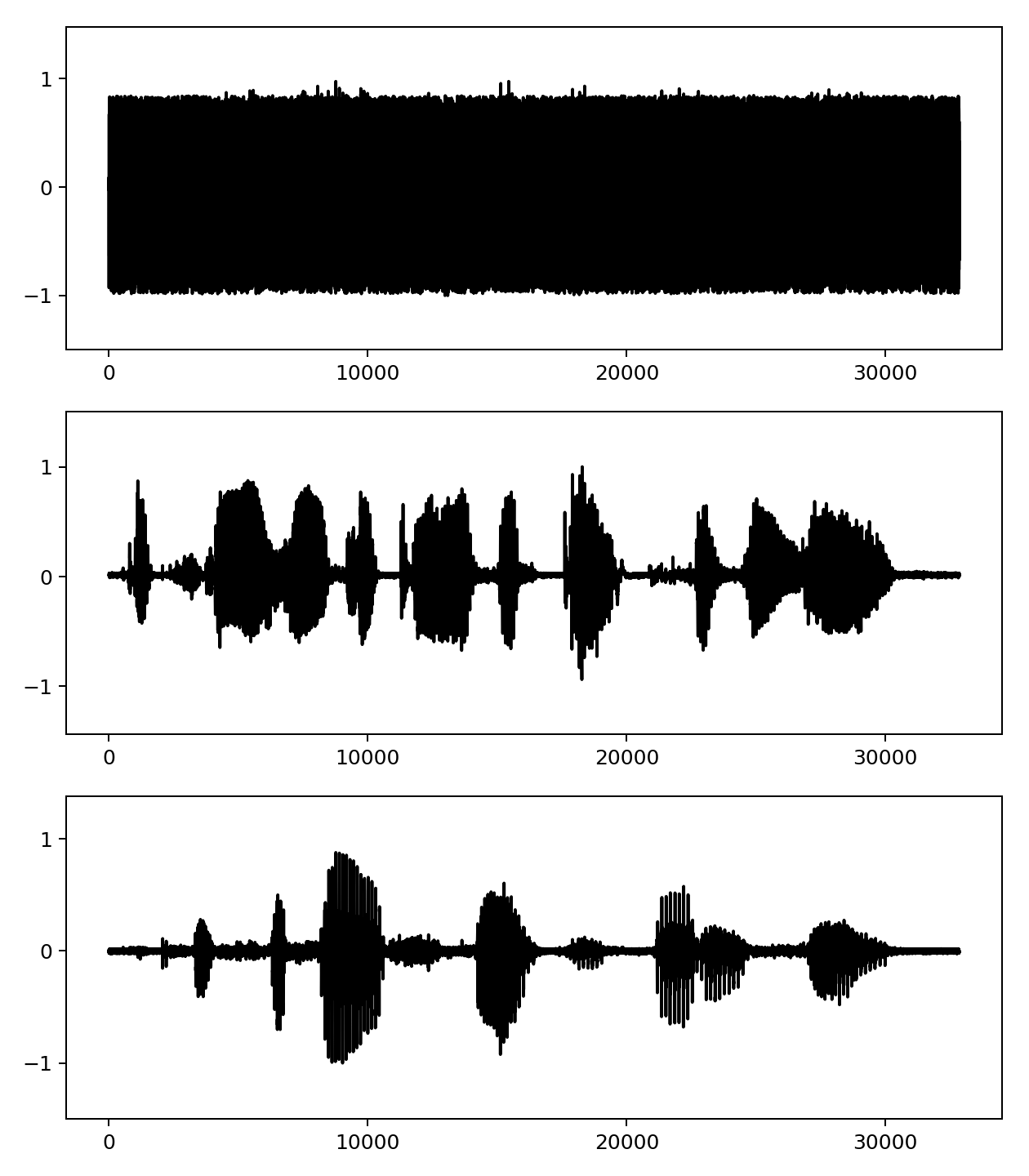}
    }
    \caption{Sources and reconstructions for the post-nonlinear mixture of audio signals.}
\end{figure}

%% file: main.bbl
\begin{thebibliography}{32}
\providecommand{\natexlab}[1]{#1}
\providecommand{\url}[1]{\texttt{#1}}
\expandafter\ifx\csname urlstyle\endcsname\relax
  \providecommand{\doi}[1]{doi: #1}\else
  \providecommand{\doi}{doi: \begingroup \urlstyle{rm}\Url}\fi

\bibitem[Almeida(2003)]{almeida2003misep}
Lu{\'\i}s~B Almeida.
\newblock Misep--linear and nonlinear ica based on mutual information.
\newblock \emph{Journal of Machine Learning Research}, 4\penalty0
  (Dec):\penalty0 1297--1318, 2003.

\bibitem[Arjovsky et~al.(2017)Arjovsky, Chintala, and
  Bottou]{arjovsky2017wasserstein}
Martin Arjovsky, Soumith Chintala, and L{\'e}on Bottou.
\newblock Wasserstein gan.
\newblock \emph{arXiv preprint arXiv:1701.07875}, 2017.

\bibitem[Bell \& Sejnowski(1995)Bell and Sejnowski]{bell1995information}
Anthony~J Bell and Terrence~J Sejnowski.
\newblock An information-maximization approach to blind separation and blind
  deconvolution.
\newblock \emph{Neural computation}, 7\penalty0 (6):\penalty0 1129--1159, 1995.

\bibitem[Chen et~al.(2016)Chen, Duan, Houthooft, Schulman, Sutskever, and
  Abbeel]{chen2016infogan}
Xi~Chen, Yan Duan, Rein Houthooft, John Schulman, Ilya Sutskever, and Pieter
  Abbeel.
\newblock Infogan: Interpretable representation learning by information
  maximizing generative adversarial nets.
\newblock \emph{CoRR}, abs/1606.03657, 2016.
\newblock URL \url{http://arxiv.org/abs/1606.03657}.

\bibitem[Comon(1994)]{comon1994independent}
Pierre Comon.
\newblock Independent component analysis, a new concept?
\newblock \emph{Signal processing}, 36\penalty0 (3):\penalty0 287--314, 1994.

\bibitem[Dinh et~al.(2014)Dinh, Krueger, and Bengio]{dinh2014nice}
Laurent Dinh, David Krueger, and Yoshua Bengio.
\newblock Nice: Non-linear independent components estimation.
\newblock \emph{arXiv preprint arXiv:1410.8516}, 2014.

\bibitem[Dinh et~al.(2016)Dinh, Sohl-Dickstein, and Bengio]{dinh2016density}
Laurent Dinh, Jascha Sohl-Dickstein, and Samy Bengio.
\newblock Density estimation using real nvp.
\newblock \emph{arXiv preprint arXiv:1605.08803}, 2016.

\bibitem[Ganin et~al.(2016)Ganin, Ustinova, Ajakan, Germain, Larochelle,
  Laviolette, Marchand, and Lempitsky]{ganin2016domain}
Yaroslav Ganin, Evgeniya Ustinova, Hana Ajakan, Pascal Germain, Hugo
  Larochelle, Fran{\c{c}}ois Laviolette, Mario Marchand, and Victor Lempitsky.
\newblock Domain-adversarial training of neural networks.
\newblock \emph{Journal of Machine Learning Research}, 17\penalty0
  (59):\penalty0 1--35, 2016.

\bibitem[Goodfellow et~al.(2014)Goodfellow, Pouget-Abadie, Mirza, Xu,
  Warde-Farley, Ozair, Courville, and Bengio]{goodfellow2014gan}
Ian Goodfellow, Jean Pouget-Abadie, Mehdi Mirza, Bing Xu, David Warde-Farley,
  Sherjil Ozair, Aaron Courville, and Yoshua Bengio.
\newblock Generative adversarial nets.
\newblock In \emph{Advances in neural information processing systems}, pp.\
  2672--2680, 2014.

\bibitem[Gretton et~al.(2005)Gretton, Herbrich, Smola, Bousquet, and
  Sch{\"o}lkopf]{gretton2005kernel}
Arthur Gretton, Ralf Herbrich, Alexander Smola, Olivier Bousquet, and Bernhard
  Sch{\"o}lkopf.
\newblock Kernel methods for measuring independence.
\newblock \emph{Journal of Machine Learning Research}, 6\penalty0
  (Dec):\penalty0 2075--2129, 2005.

\bibitem[Gretton et~al.(2008)Gretton, Fukumizu, Teo, Song, Sch{\"o}lkopf, and
  Smola]{gretton2008kernel}
Arthur Gretton, Kenji Fukumizu, Choon~H Teo, Le~Song, Bernhard Sch{\"o}lkopf,
  and Alex~J Smola.
\newblock A kernel statistical test of independence.
\newblock In \emph{Advances in neural information processing systems}, pp.\
  585--592, 2008.

\bibitem[Gulrajani et~al.(2017)Gulrajani, Ahmed, Arjovsky, Dumoulin, and
  Courville]{gulrajani2017improved}
Ishaan Gulrajani, Faruk Ahmed, Martin Arjovsky, Vincent Dumoulin, and Aaron
  Courville.
\newblock Improved training of wasserstein gans.
\newblock \emph{arXiv preprint arXiv:1704.00028}, 2017.

\bibitem[Hjelm et~al.(2017)Hjelm, Jacob, Che, Cho, and Bengio]{hjelm2017}
R~Devon Hjelm, Athul~Paul Jacob, Tong Che, Kyunghyun Cho, and Yoshua Bengio.
\newblock Boundary-seeking generative adversarial networks.
\newblock \emph{arXiv preprint arXiv:1702.08431}, 2017.

\bibitem[Huszar(2016)]{huszar2016alternative}
Ferenc Huszar.
\newblock An alternative update rule for generative adversarial networks.
\newblock \emph{Unpublished note (retrieved on 7 Oct 2016)}, 2016.

\bibitem[Hyvarinen \& Morioka(2017)Hyvarinen and
  Morioka]{hyvarinen2017nonlinear}
Aapo Hyvarinen and Hiroshi Morioka.
\newblock Nonlinear ica of temporally dependent stationary sources.
\newblock In \emph{Artificial Intelligence and Statistics}, pp.\  460--469,
  2017.

\bibitem[Hyv{\"a}rinen \& Oja(1997)Hyv{\"a}rinen and Oja]{hyvarinen1997one}
Aapo Hyv{\"a}rinen and Erkki Oja.
\newblock One-unit learning rules for independent component analysis.
\newblock \emph{Advances in neural information processing systems}, pp.\
  480--486, 1997.

\bibitem[Hyv\"{a}rinen et~al.(2004)Hyv\"{a}rinen, Karhunen, and
  Oja]{hyvarinen2004ica}
Aapo Hyv\"{a}rinen, Juha Karhunen, and Erkki Oja.
\newblock \emph{Independent component analysis}, volume~46.
\newblock John Wiley \& Sons, 2004.

\bibitem[Ioffe \& Szegedy(2015)Ioffe and Szegedy]{ioffe2015batch}
Sergey Ioffe and Christian Szegedy.
\newblock Batch normalization: Accelerating deep network training by reducing
  internal covariate shift.
\newblock \emph{arXiv preprint arXiv:1502.03167}, 2015.

\bibitem[Kabal(2002)]{kabal2002tsp}
Peter Kabal.
\newblock Tsp speech database.
\newblock \emph{McGill University, Database Version}, 1\penalty0 (0):\penalty0
  09--02, 2002.

\bibitem[Kingma \& Welling(2013)Kingma and Welling]{kingma2013auto}
Diederik~P Kingma and Max Welling.
\newblock Auto-encoding variational bayes.
\newblock \emph{arXiv preprint arXiv:1312.6114}, 2013.

\bibitem[Kraskov et~al.(2004)Kraskov, St{\"o}gbauer, and
  Grassberger]{kraskov2004estimating}
Alexander Kraskov, Harald St{\"o}gbauer, and Peter Grassberger.
\newblock Estimating mutual information.
\newblock \emph{Physical review E}, 69\penalty0 (6):\penalty0 066138, 2004.

\bibitem[Makhzani et~al.(2015)Makhzani, Shlens, Jaitly, Goodfellow, and
  Frey]{makhzani2015adversarial}
Alireza Makhzani, Jonathon Shlens, Navdeep Jaitly, Ian Goodfellow, and Brendan
  Frey.
\newblock Adversarial autoencoders.
\newblock \emph{arXiv preprint arXiv:1511.05644}, 2015.

\bibitem[Mao et~al.(2016)Mao, Li, Xie, Lau, Wang, and Smolley]{mao2016least}
Xudong Mao, Qing Li, Haoran Xie, Raymond~YK Lau, Zhen Wang, and Stephen~Paul
  Smolley.
\newblock Least squares generative adversarial networks.
\newblock \emph{arXiv preprint ArXiv:1611.04076}, 2016.

\bibitem[Naik \& Kumar(2011)Naik and Kumar]{naik2011overview}
Ganesh~R Naik and Dinesh~K Kumar.
\newblock An overview of independent component analysis and its applications.
\newblock \emph{Informatica}, 35\penalty0 (1), 2011.

\bibitem[Nair \& Hinton(2010)Nair and Hinton]{nair2010rectified}
Vinod Nair and Geoffrey~E Hinton.
\newblock Rectified linear units improve restricted boltzmann machines.
\newblock In \emph{Proceedings of the 27th international conference on machine
  learning (ICML-10)}, pp.\  807--814, 2010.

\bibitem[Pedregosa et~al.(2011)Pedregosa, Varoquaux, Gramfort, Michel, Thirion,
  Grisel, Blondel, Prettenhofer, Weiss, Dubourg, et~al.]{pedregosa2011scikit}
Fabian Pedregosa, Ga{\"e}l Varoquaux, Alexandre Gramfort, Vincent Michel,
  Bertrand Thirion, Olivier Grisel, Mathieu Blondel, Peter Prettenhofer, Ron
  Weiss, Vincent Dubourg, et~al.
\newblock Scikit-learn: Machine learning in python.
\newblock \emph{Journal of Machine Learning Research}, 12\penalty0
  (Oct):\penalty0 2825--2830, 2011.

\bibitem[Schmidhuber(1992)]{schmidhuber1992learning}
J{\"u}rgen Schmidhuber.
\newblock Learning factorial codes by predictability minimization.
\newblock \emph{Neural Computation}, 4\penalty0 (6):\penalty0 863--879, 1992.

\bibitem[Serdyuk et~al.(2016)Serdyuk, Audhkhasi, Brakel, Ramabhadran, Thomas,
  and Bengio]{serdyuk2016invariant}
Dmitriy Serdyuk, Kartik Audhkhasi, Phil{\'e}mon Brakel, Bhuvana Ramabhadran,
  Samuel Thomas, and Yoshua Bengio.
\newblock Invariant representations for noisy speech recognition.
\newblock \emph{arXiv preprint arXiv:1612.01928}, 2016.

\bibitem[St{\"o}gbauer et~al.(2004)St{\"o}gbauer, Kraskov, Astakhov, and
  Grassberger]{stogbauer2004least}
Harald St{\"o}gbauer, Alexander Kraskov, Sergey~A Astakhov, and Peter
  Grassberger.
\newblock Least-dependent-component analysis based on mutual information.
\newblock \emph{Physical Review E}, 70\penalty0 (6):\penalty0 066123, 2004.

\bibitem[Taleb \& Jutten(1999)Taleb and Jutten]{taleb1999source}
Anisse Taleb and Christian Jutten.
\newblock Source separation in post-nonlinear mixtures.
\newblock \emph{IEEE transactions on Signal Processing}, 47\penalty0
  (10):\penalty0 2807--2820, 1999.

\bibitem[Tieleman \& Hinton(2012)Tieleman and Hinton]{tieleman2012lecture}
Tijmen Tieleman and Geoffrey Hinton.
\newblock Lecture 6.5-rmsprop: Divide the gradient by a running average of its
  recent magnitude.
\newblock \emph{COURSERA: Neural networks for machine learning}, 4\penalty0
  (2), 2012.

\bibitem[Zheng et~al.(2007)Zheng, Huang, Li, Irwin, and Sun]{zheng2007misep}
Chun-Hou Zheng, De-Shuang Huang, Kang Li, George Irwin, and Zhan-Li Sun.
\newblock Misep method for postnonlinear blind source separation.
\newblock \emph{Neural computation}, 19\penalty0 (9):\penalty0 2557--2578,
  2007.

\end{thebibliography}
